\documentclass[letterpaper,journal]{IEEEtran}
\usepackage{amsmath,amsfonts,amssymb,amsthm}
\usepackage{mathtools}
\usepackage{algorithmic}
\usepackage{algorithm}
\usepackage{array}
\usepackage{textcomp}
\usepackage{stfloats}
\usepackage{url}
\usepackage{verbatim}
\usepackage{graphicx}
\usepackage{pgfplots}
\usepackage{cite}
\usepackage{microtype}
\usepackage{booktabs}
\usepackage{balance}
\usepackage{tabularx}
\usepackage{multirow}
\usepackage{colortbl}
\usepackage{xcolor}
\usepackage[hidelinks]{hyperref}

\pgfplotsset{compat=1.18}

\newcommand{\cmark}{\checkmark}
\newcommand{\xmark}{$\times$}
\newcommand{\model}{\textsc{MASIRL}}

\theoremstyle{plain}
\newtheorem{theorem}{Theorem}

\theoremstyle{definition}

\theoremstyle{remark}
\newtheorem{remark}[theorem]{Remark}

\begin{document}

\title{Multi-Agent Self-Improving Reinforcement Learning for Video Reasoning}

\author{%
	Mingwen Zhang,
	Jisheng Dang,
	Minqiang Yang,
    Bimei Wang,
	Bin Hu,~\IEEEmembership{Fellow, ~IEEE},
	Tat-Seng Chua%
    \thanks{This work was supported in part by the National Natural Science Foundation of China (Grant No.62472203, No.62227807 and Grant No.U24B20186), in part by the STI 2030-Major Projects (2021ZD0202002), in part by the Fundamental Research Funds for the Central Universities (Grant No. lzujbky-2023-16), in part by the Gansu Provincial Science and Technology Program (Grant No. 26JRRA258).This work was also supported by the Talent Scientific Fund of Lanzhou University and in part by Supercomputing Center of Lanzhou University.}
    \thanks{Mingwen Zhang, Jisheng Dang, Minqiang Yang and Bimei Wang are with the School of information Science and Engineering, Lanzhou University, LanZhou, China (E-mail: zhangmw2024@lzu.edu.cn, 
    dangjsh@mail2.sysu.edu.cn, yangmq@lzu.edu.cn, 
    wangbimei@lzu.edu.cn).}
	\thanks{Bin Hu is with the Gansu Provincial Key Laboratory of Wearable Computing, School of Information Science and Engineering, Lanzhou University, Lanzhou, China, and with the Brain Health Engineering Laboratory, Institute  of Engineering Medicine, Beijing Institute of Technology, Beijing, China(E-mail: bh@lzu.edu.cn). }
    \thanks{Tat-Seng Chua is with the School of Computing, National University of Singapore, Singapore 119077 (E-mail: dcscts@nus.edu.sg). }
    \thanks{*Corresponding author: Jisheng Dang, Minqiang Yang, and Bin Hu.}
}

\maketitle

\begin{abstract}
	Video reasoning tasks such as grounded video question answering and temporal grounding require selecting temporal evidence that supports the query. In many current training setups, temporal supervision is applied through local objectives such as boundary regression or span generation, while verification is used mainly to rerank candidate segments at inference time. We study whether a frozen verifier can also guide training. Our multi-agent framework couples a trainable \emph{Grounder} with a frozen \emph{Verifier}: the Grounder samples candidate trajectories and evidence segments, the Verifier assigns query-conditioned segment scores, a group-relative policy-gradient objective favors trajectories that outperform their within-input peers, and a bootstrapped calibration loss steers temporal predictions toward verifier-preferred spans. Trained on source tasks and evaluated without target-dataset fine-tuning, a two-billion-parameter instantiation transfers zero-shot across grounded question answering, temporal grounding, and long-video question answering, reaching 28.7\% intersection-over-union and 25.4\% answer-grounding accuracy on a grounded-question-answering benchmark, 46.1\% intersection-over-union on a temporal-grounding benchmark, and 54.1\% on a long-video question-answering benchmark. Relative to a strong same-scale baseline, the gains are modest but consistent, with the clearest improvements on relevance-oriented metrics such as intersection-over-union and moderate-overlap recall. Within the tested benchmarks and transfer setting, the results support frozen verification as a training signal for evidence selection, while showing that strict boundary precision remains comparatively weaker. Code and models are available at \url{https://anonymous.4open.science/r/MASIRL-E50C/}.
\end{abstract}


\begin{figure}[!t]
	\centering
	\includegraphics[width=\columnwidth]{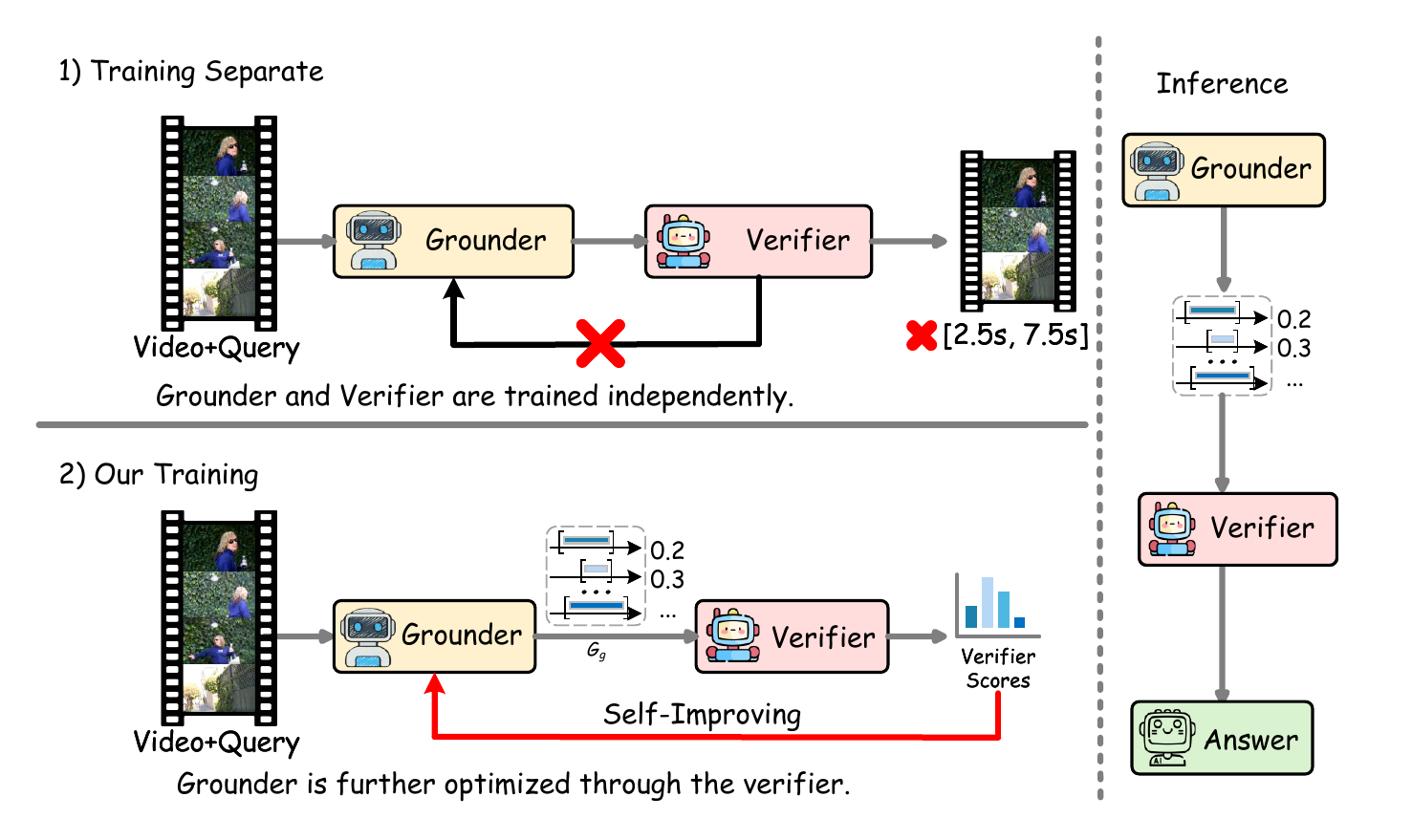}
	\caption{Comparison of two training paradigms for video reasoning.
		\textbf{Top:} Separate training, in which the Grounder and the Verifier are optimized independently and interact only during inference-time candidate selection.
		\textbf{Right:} The inference flow shared by both paradigms, from the input query through candidate generation, verification, and final selection.
		\textbf{Bottom:} Training-time integration in our framework, in which the frozen Verifier scores candidate evidence segments and the Grounder is updated from those scores.
		Gradients propagate only through the Grounder.}
	\label{fig:intro}
\end{figure}

\section{Introduction}
\label{sec:intro}
\IEEEPARstart{V}{ideo} reasoning tasks such as grounded video question answering and temporal grounding demand more than surface-form generation: they require identifying a temporal interval whose content genuinely supports a given query~\cite{zhang2023temporal}.
Although recent video-language models can produce answers and temporal spans through a unified generative interface~\cite{li2025universal,wang2024grounded}, their practical value still depends on whether the predicted interval captures the event that justifies the answer while excluding semantically related distractors.
For this paper, we therefore treat video reasoning as a video-evidence selection problem: the model must not only answer the query, but also identify the segment that best supports that answer.
This requirement matters to TIP readers because localization quality directly affects grounded interpretation, retrieval reliability, and downstream reasoning over long, cluttered video streams.

Current training signals only partially address that requirement.
In many grounding-oriented pipelines, supervision is still expressed through endpoint regression, span generation, or proposal scoring~\cite{yuan2021closer,kang2025empower}, which encourages local boundary decisions but gives limited direct preference among several semantically plausible intervals for the same query.
Separately, verifier or reranking modules are often introduced after candidate generation to improve final selection, so their judgments need not shape the policy that proposes evidence during training.
This mismatch leaves a practical gap: models can reach the right semantic neighborhood while still preferring segments that are too loose, too inclusive, or weakly justified.
Sequence-level reinforcement learning is a natural way to optimize interval quality beyond token-level likelihood or local boundary losses.
Yet in long-video settings the rewards are sparse, near-miss segments are common, and stochastic decoding complicates credit assignment~\cite{ouyang2022instructgpt,li2025reinforcement,li2025videochat}.
What is missing is a training signal that scores candidate evidence segments at the sequence level while remaining stable enough to guide optimization.

We address that gap by moving verification from a post-hoc selector into the training loop.
Specifically, we train a verifier on source verification data, freeze it during Grounder optimization, and use its query-conditioned segment scores to evaluate candidate evidence spans.
Because the verifier is frozen, reward construction remains decoupled from Grounder updates; because it scores explicit segments rather than free-form outputs, the signal stays tied to evidence quality in the paper's target setting.

Building on this design, we introduce \model, a multi-agent self-improving reinforcement learning framework organized around two complementary roles.
A trainable \emph{Grounder} samples answer-bearing trajectories and proposes candidate evidence segments, whereas a frozen \emph{Verifier} scores those candidates throughout training.
The resulting Verifier scores serve a dual purpose: they drive a group-relative REINFORCE (GRR) objective that promotes trajectories outperforming their within-input peers, and they supply pseudo-targets for an auxiliary calibration loss that steers Grounder's temporal head toward verifier-preferred spans.
Verifier scores are further employed to gate updates, thereby down-weighting trajectories that the Verifier already judges as strong.
Figure~\ref{fig:intro} illustrates this interaction; gradients flow exclusively through the Grounder.

Following supervised source-task training and without any target-dataset fine-tuning, a 2B \model\ transfers zero-shot to grounded QA, temporal grounding, and long-video QA.
On NExT-GQA the model attains 28.7 mIoU and 25.4\% Acc@GQA; on Charades-STA it reaches 46.1 mIoU; and on Video-MME it achieves 54.1\%.
Across these benchmarks, \model\ improves over the strongest same-scale 2B baseline, while larger or differently trained systems remain contextual rather than primary comparisons.
The improvements are modest but directionally consistent, and they are clearest on relevance-oriented metrics such as mIoU, R@0.3, and R@0.5 rather than on the strictest boundary thresholds.
Within this zero-shot transfer setting, the paper's contribution is therefore bounded: it shows that frozen verification can serve as a training signal for video evidence selection, while strict boundary precision remains a weaker aspect of the current design.

The contributions of this work are threefold:
\begin{itemize}
	\item We formulate multi-agent training for video reasoning by using a frozen verifier to provide query-conditioned segment scores during Grounder optimization, instead of limiting verification to inference-time reranking.
	\item We develop a training objective that combines group-relative REINFORCE, bootstrapped boundary calibration, and verifier-based update gating so that sequence-level policy updates and temporal refinement are learned within one framework.
	\item We show that, after source-task training, a 2B model transfers zero-shot across grounded QA, temporal grounding, and long-video QA without target-dataset fine-tuning, with the most consistent gains appearing on relevance-oriented metrics rather than strict boundary thresholds.
\end{itemize}

\section{Related Work}
\label{sec:relatedwork}

\begin{figure*}[!t]
	\centering
	\includegraphics[width=0.95\textwidth]{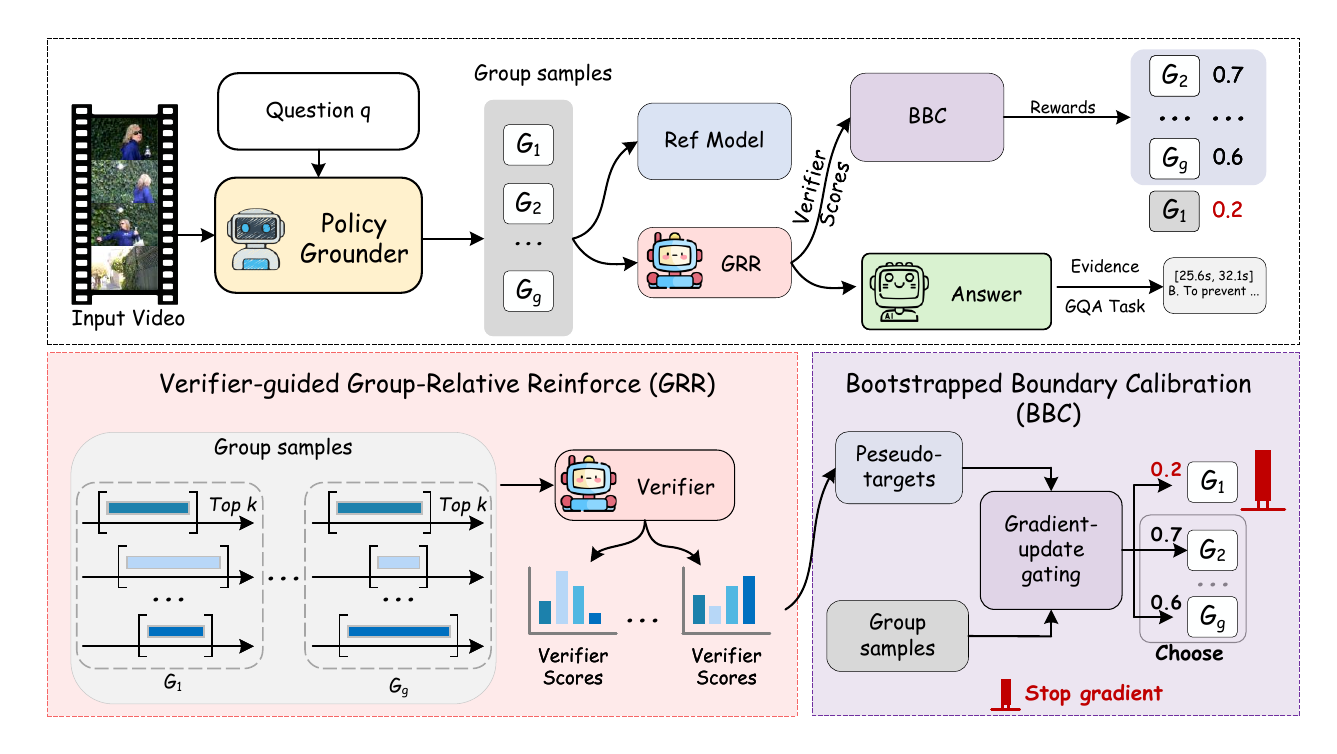}
	\caption{Overview of the Grounder--Verifier training framework used in this paper.
		The Grounder generates candidate temporal segments from video and query inputs.
		GRR samples multiple trajectories and updates the policy from within-group relative advantages.
		Bootstrapped boundary calibration uses Verifier-selected pseudo-targets to refine temporal predictions.
		Gradients propagate only through the Grounder, while the Verifier remains frozen and decoupled from optimization.}
	\label{fig:framework}
\end{figure*}

\noindent \textbf{Temporal Sentence Grounding in Videos.}
Temporal sentence grounding in videos (TSGV), also referred to as moment retrieval, seeks to localize the interval within an untrimmed video that corresponds to a natural-language query.
The task was first formalized through dedicated datasets and retrieval-style formulations that align queries with candidate moments~\cite{hendricks2017localizing,gao2017tall}.
Subsequent methods branched into proposal-based, matching-based, and span-based paradigms, each combining candidate generation, cross-modal alignment, and boundary regression in different configurations~\cite{liu2018attentive,liu2018crossmodal,chen2019semantic,zhang2019cmin,zhang2020learning,zhang2020span,yuan2021scdm_tip}.
These works are the closest training precedents for our setting because they optimize where a segment starts and ends, but they usually do so through local proposal scores or endpoint losses.
That design differs from our focus: we optimize how candidate evidence segments are ranked against one another during training, not only how a single predicted span matches a target boundary.
Later datasets and evaluation protocols made this distinction more visible by exposing the gap between semantic relevance and exact boundary precision~\cite{lei2021detecting,yuan2021closer,li2024unsupervised_highlight_tip}.

Recent generative video-language models push grounding into broader instruction-following settings~\cite{li2025universal,wang2024grounded,kang2025empower,wang2025mllmta_tip}.
They are nearest to our work at the model-interface level because they jointly produce answers and temporal spans, but their supervision is still largely anchored in span prediction or supervised generation.
Our point of departure is not the use of a generative backbone itself, but the introduction of a frozen verifier as a training-time signal for choosing among candidate evidence segments.
Graph-based and compositional video question answering methods~\cite{cheng2024keyword_tip,yuan2024eventgraph_tip,xiao2024dynamic_stgraph_tip} provide richer reasoning structures, yet they target representation and reasoning design rather than multi-agent evidence selection.

\noindent \textbf{Reinforcement Learning and Group-Relative Policy Optimization.}
Reinforcement learning provides a principled mechanism for optimizing non-differentiable localization criteria such as IoU, casting temporal grounding as a sequential decision problem~\cite{he2019readwatchmove}.
In the broader context of sequence prediction, self-critical training and related policy-gradient baselines reduce variance by contrasting a sampled trajectory against a reference~\cite{rennie2017self}.
Preference-based and post-training pipelines have further underscored the importance of reward design, baseline selection, and stable policy improvement in settings where the objective is only loosely aligned with token-level likelihood~\cite{schulman2017ppo,ouyang2022instructgpt,rafailov2023dpo}.
Among more recent developments, group-based objectives such as GRPO draw multiple samples per input and derive relative advantages within the group, thereby improving stability without requiring a learned critic~\cite{shao2024deepseekmath}.
This family of methods is the nearest optimization analogue for our approach because video reasoning often presents several plausible trajectories for the same query.
Our GRR objective follows that group-relative view, but adapts it to candidate evidence selection in video by combining overlap-based supervision, when labels are available, with verifier-based segment scores.
The resulting difference is narrow but important: the objective is tuned to ranking evidence-bearing trajectories in multimodal grounding, not to generic language-model post-training.

\noindent \textbf{Multi-Agent Learning and Verification.}
Verification models and reward models have become central to recent sequence-level post-training.
Preference-based RLHF decouples reward construction from policy updates~\cite{ouyang2022instructgpt,christiano2017preferences,schulman2017ppo,rafailov2023dpo}, process-supervised verification supplies step-level feedback for reasoners~\cite{lightman2024verify}, and self-improving pipelines bootstrap from model-generated targets~\cite{zelikman2022star,hosseini2024vstar}.
These studies establish the broad idea that learned evaluators can guide training rather than act only as inference-time filters.
Our nearest point of contact is that reuse of evaluator signals during optimization, but the present setting is narrower: the verifier scores explicit video segments for evidence relevance, not free-form textual solutions or reasoning traces.
In a parallel thread, multi-agent systems assign specialized roles for planning, retrieval, and validation~\cite{du2024multiagent,rashid2018qmix}, although that literature is not mainly concerned with temporal evidence localization in video.
Relative to both multi-agent post-training and multi-agent coordination, our delta is twofold.
First, the Verifier remains frozen throughout Grounder optimization, so it serves as a fixed segment scorer rather than a jointly updated critic or reward model.
Second, the verifier's role is deliberately task-specific: it scores candidate evidence spans for temporal grounding and grounded reasoning, rather than coordinating a general reasoning pipeline.
These choices position the method as multi-agent evidence selection for video reasoning, not as a broad RLHF substitute or a general multi-agent architecture.
Accordingly, the paper does not claim a general multi-agent reasoning framework; rather, it studies how a frozen verifier can be reused as a training-time signal for video evidence selection.

\section{Method}
\label{sec:method}

\noindent This section defines how \model uses a frozen Verifier to train a Grounder without changing the underlying formulation in Figure~\ref{fig:framework}.
For each video-query pair $x=(v,q)$, the Grounder samples a group of trajectories, each of which proposes one or more temporal evidence segments.
The frozen Verifier then scores these candidate segments conditioned on the same query, and those scores are reused to define trajectory rewards, select pseudo-targets for boundary calibration, and gate which groups receive updates.
Gradients propagate only through the Grounder; the Verifier acts as a fixed segment scorer throughout training.
The remainder of this section specifies the Grounder--Verifier setup (Sec.~\ref{subsec:setup}), the multi-agent group-relative REINFORCE objective (Sec.~\ref{subsec:grr}), the bootstrapped boundary calibration mechanism (Sec.~\ref{subsec:calib}), the Verifier-based update gating rule (Sec.~\ref{subsec:gating}), and the combined training objective (Sec.~\ref{subsec:objective}).

\subsection{Grounder and Verifier Setup}
\label{subsec:setup}

This subsection defines the two agent roles and the candidate representation passed between them.
The framework adopts a dual-agent architecture comprising a \textbf{Grounder} agent responsible for generation and localization, and a \textbf{Verifier} agent responsible for scoring candidate evidence segments.
Let $x = (v, q)$ denote video $v$ and language query $q$.
The video is discretized into $T$ clips indexed by $t \in \{0,\dots,T-1\}$, and both video and query are encoded into a joint sequence of representations.
The \textbf{Grounder} $\pi_\theta$ is a generative multimodal model that, conditioned on $x$, produces a completion $c \sim \pi_\theta(\cdot \mid x)$ and, within the same forward pass, predicts temporal boundaries through an auxiliary temporal head attached to the joint video-query representations.
Together, this temporal head and the shared backbone yield one or more candidate intervals $\mathcal{C}_\theta(x, c) = \{y_k\}_{k=1}^K$ with $y_k = [s_k, e_k]$ along the video timeline.
The head may operate in a \emph{discrete} mode, producing start/end distributions over clip indices, or in a \emph{continuous} mode, mapping normalized coordinates onto the timeline; in either case, invalid or duplicate spans are filtered and the retained candidates are converted to a common temporal coordinate system.
This candidate-formation stage exposes several plausible evidence intervals to the Verifier rather than forcing the Grounder to commit to a single span before sequence-level feedback becomes available.

The frozen \textbf{Verifier} $V_\phi$ scores a candidate segment $y$ \emph{conditioned on $q$ and on the video content restricted to $y$}.
Concretely, scoring is based on a log-likelihood ratio
$\Delta_\phi(x,y) = \log p_\phi(\texttt{Yes.}\mid x,y) - \log p_\phi(\texttt{No.}\mid x,y)$ under teacher forcing, from which we define $V_\phi(x,y) = \sigma(\Delta_\phi(x,y)) \in [0,1]$, with $\sigma$ denoting the logistic sigmoid.
Formulating the score as a log-likelihood ratio, rather than relying on a single decision token, enables a direct comparison of positive and negative hypotheses.
The Verifier is kept frozen and participates only in scoring and selection; no gradient propagates through $V_\phi$ during training.
Segment boundaries are explicitly marked in the multimodal input so that the Verifier attends to a specified interval rather than to the full video.
Trained separately on video, query, and segment triplets, the Verifier is optimized for \emph{segment-level relevance} rather than for strict IoU or R@0.7.
In total, 232K verification triplets are constructed from DiDeMo-Verify, TACoS-Verify, and QVHighlights-Verify (official training splits only), all disjoint from the evaluation benchmarks; a complete role-wise data configuration appears in Table~\ref{tab:role_datasets}.
Each verification triplet comprises a query, a source video, and an explicitly marked candidate interval, ensuring that the decision concerns \emph{whether a given segment constitutes relevant evidence} rather than whether the video as a whole is pertinent.
Casting the score as a Yes/No log-likelihood difference is more stable than reading a single affirmative token probability, since both hypotheses are evaluated under the same prompt and segment context.
The triplets encompass positive evidence segments alongside challenging negatives derived from mismatched or temporally off-target intervals, so that the Verifier learns to favor semantically correct evidence while remaining tolerant of moderate boundary perturbations.

\subsection{Multi-Agent Group-Relative REINFORCE}
\label{subsec:grr}

This subsection defines how verifier scores become sequence-level rewards for Grounder updates.
For each input $x$, the Grounder draws $G$ completions $c_1, \ldots, c_G \sim \pi_\theta(\cdot \mid x)$.
Each completion $c_g$ gives rise to a candidate set $\mathcal{C}_\theta(x, c_g)$ produced by the shared temporal head.
To assign a scalar reward to each trajectory, a single \emph{reward-carrier} segment $y_g^{(r)} \in \mathcal{C}_\theta(x, c_g)$, representing the trajectory's strongest piece of evidence, is first identified.

\paragraph{Reward Definition.}
The quality of the reward carrier is quantified as
\begin{equation}
	q_g = \alpha\,\mathrm{IoU}(y_g^{(r)}, y^\star) + \beta_V\,V_\phi(x, y_g^{(r)}),
	\label{eq:q_reward}
\end{equation}
and the trajectory reward is
\begin{equation}
	r_g = \gamma\,\mathbb{I}_{\mathrm{reg}}(c_g) + \mathbb{I}_{\mathrm{valid}}(c_g)\,q_g,
	\label{eq:reward}
\end{equation}
where $y^\star$ denotes the ground-truth interval when available and $\mathrm{IoU}$ is the 1D intersection-over-union.
The indicator $\mathbb{I}_{\mathrm{reg}}$ penalizes malformed outputs, whereas $\mathbb{I}_{\mathrm{valid}}$ gates the quality score on structurally valid completions.
In the absence of annotations the IoU term is dropped ($\alpha=0$), leaving the Verifier as sole source of quality signal.
This decomposition provides a common objective across fully supervised, weakly supervised, and zero-shot regimes.

\paragraph{Reward-carrier Selection.}
The reward carrier $y_g^{(r)}$ is the single segment through which a trajectory receives sequence-level credit.
Under the default configuration, the Verifier ranks candidates within $\mathcal{C}_\theta(x,c_g)$, retains a short Top-$K_r$ list, and designates the highest-scoring valid span as $y_g^{(r)}$.
Restricting credit to one representative segment keeps the scalar reward tied to a single evidence choice and avoids averaging across heterogeneous segments within the same trajectory.

\paragraph{Group-relative Advantage and GRR Loss.}
A leave-one-out baseline $b_g = \frac{1}{G-1} \sum_{j \neq g} r_j$ yields within-group advantages $A_g = r_g - b_g$, ensuring that updates reflect \emph{relative} quality among sampled trajectories rather than absolute reward magnitudes.
Inspired by GRPO~\cite{shao2024deepseekmath}, this group-relative formulation is adapted here to video reasoning, where rewards combine localization overlap with Verifier scores.
Letting $\mathrm{NLL}_\theta(x, c_g)$ denote the token-averaged negative log-likelihood of $c_g$ under $\pi_\theta$ (with prompt tokens masked), the GRR loss to be minimized is
\begin{equation}
	\mathcal{L}_{\mathrm{GRR}}(\theta) = \frac{1}{G} \sum_{g=1}^{G} A_g \cdot \mathrm{NLL}_\theta(x, c_g).
	\label{eq:grr_loss}
\end{equation}
Trajectories for which $A_g > 0$ are reinforced through a reduction in NLL, whereas those with $A_g < 0$ are suppressed.
Differentiating \eqref{eq:grr_loss} recovers the standard REINFORCE estimator equipped with a leave-one-out baseline:
\begin{equation}
	-\nabla_\theta \mathcal{L}_{\mathrm{GRR}}(\theta)
	= \mathbb{E}_{c_g \sim \pi_\theta(\cdot \mid x)}\bigl[A_g \,\nabla_\theta \log \pi_\theta(c_g \mid x)\bigr].
	\label{eq:reinforce}
\end{equation}
Because $b_g$ depends solely on rewards from other trajectories within the same group, the estimator remains unbiased, while the group-relative normalization reduces variance and renders updates less sensitive to the global reward scale.

\begin{algorithm}[!t]
	\caption{Verifier-guided training procedure with GRR, calibration, and gating}
	\label{alg:training}
	\begin{algorithmic}[1]
		\REQUIRE Video-query pairs $(v, q)$, Grounder $\pi_\theta$, frozen Verifier $V_\phi$
		\REQUIRE Hyperparameters: group size $G$, calibration pool size $K_{\mathrm{cal}}$, temperature $\tau$, loss weight $\lambda$, gating ratio $\rho$, window $K$, warmup $T_{\mathrm{warmup}}$
		\FOR{each epoch $t = 1, 2, \ldots$}
		\STATE Evaluate mean Verifier score $m_t$ on validation candidates
		\IF{$t > T_{\mathrm{warmup}}$}
		\STATE $\tau_t \gets \rho \cdot \frac{1}{\min(t,K)}\sum_{t'=t-\min(t,K)+1}^{t} m_{t'}$
		\ENDIF
		\FOR{each batch $(v, q)$}
		\STATE Sample $G$ completions $c_1, \ldots, c_G \sim \pi_\theta(\cdot \mid (v, q))$
		\STATE For each $g$, parse $\mathcal{C}_\theta((v,q), c_g)$, choose $y_g^{(r)}$ from the Verifier-ranked Top-$K_r$ candidates, and compute $r_g$ by Eq.~\eqref{eq:reward}
		\STATE Form leave-one-out baselines $b_g = \frac{1}{G-1} \sum_{j \neq g} r_j$ and advantages $A_g = r_g - b_g$ for all $g$
		\IF{$t > T_{\mathrm{warmup}}$}
		\STATE For each $g$, let $\bar{v}_g = \max_{y \in \mathcal{C}_\theta((v,q),c_g)} V_\phi((v,q),y)$ and set $A_g \gets 0$ if $\bar{v}_g \ge \tau_t$
		\ENDIF
		\STATE Compute $\mathcal{L}_{\mathrm{GRR}} = \frac{1}{G}\sum_{g=1}^{G} A_g \cdot \mathrm{NLL}_\theta((v,q), c_g)$
		\STATE $\mathcal{T}(x) \gets \mathrm{Top}\text{-}K_{\mathrm{cal}}\bigl(\bigcup_g \mathcal{C}_\theta((v,q),c_g),\; V_\phi\bigr)$
		\STATE For $y \in \mathcal{T}(x)$, compute $w_y \propto \exp(V_\phi((v,q),y)/\tau)$ and stop gradients through $\mathcal{T}(x)$, $\{w_y\}$, and $V_\phi$
		\STATE $\mathcal{L}_{\mathrm{cal}} = \sum_{y \in \mathcal{T}(x)} w_y \,\ell(\hat{y}_\theta(x),\, y)$
		\STATE $\mathcal{L} \gets \mathcal{L}_{\mathrm{GRR}} + \lambda\,\mathcal{L}_{\mathrm{cal}}$
		\STATE Update $\theta$ via $\nabla_\theta \mathcal{L}$, without gradients into $\phi$
		\ENDFOR
		\ENDFOR
	\end{algorithmic}
\end{algorithm}

Algorithm~\ref{alg:training} is written at the level of groupwise transformations rather than per-candidate bookkeeping. In practice, each sampled completion first induces a trajectory-specific candidate pool, after which the frozen Verifier is reused in three non-overlapping roles: it selects the reward carrier that defines the GRR signal, ranks the pooled candidates that become calibration pseudo-targets, and determines whether an already-strong group should be removed from the policy-gradient update. This compact presentation preserves the same computation graph as the expanded implementation while making clear that only the Grounder parameters are optimized.

The resulting separation of responsibilities is important for interpreting the training dynamics. GRR compares trajectories against one another through leave-one-out advantages, the calibration term stabilizes the temporal head by averaging over several high-confidence spans, and gating prevents the relative-reward update from being dominated by groups whose best proposal already satisfies the current Verifier threshold. Read together with Secs.~\ref{subsec:grr}--\ref{subsec:gating}, the algorithm therefore summarizes a single training loop in which sequence-level credit assignment, boundary refinement, and difficulty-aware update selection are coordinated but not conflated.

\subsection{Bootstrapped Boundary Calibration}
\label{subsec:calib}

This subsection defines how the method supplies boundary supervision to the Grounder's temporal head.
Although GRR updates the language model, it does not furnish a direct regression target for temporal boundaries.
To address this limitation, we introduce an auxiliary calibration loss in which the Grounder refines its boundary predictions through Verifier-weighted pseudo-targets~\cite{zelikman2022star}.
Concretely, once the group-level GRR computation has identified which sampled trajectories should still contribute to learning, all candidate spans produced within that batch are merged into a shared calibration pool. The temporal head is then trained against a small Verifier-ranked subset of that pool, so the auxiliary supervision remains synchronized with the same candidate distribution that generated the policy-gradient signal, yet is not restricted to the single reward carrier used for GRR.
All candidates across the $G$ trajectories are first pooled into a single set
\begin{equation}
	\mathcal{C}_\theta(x) = \bigcup_{g=1}^{G} \mathcal{C}_\theta(x, c_g),
	\label{eq:pseudo}
\end{equation}
from which the Top-$K_{\mathrm{cal}}$ elements ranked by Verifier score are selected:
\begin{equation}
	\mathcal{T}(x) = \operatorname{Top\text{-}}K_{\mathrm{cal}}\bigl\{ y \in \mathcal{C}_\theta(x) \,\big|\, V_\phi(x,y) \bigr\},
\end{equation}
where $K_{\mathrm{cal}} \ge 1$; when fewer candidates exist, $\mathcal{T}(x) = \mathcal{C}_\theta(x)$ is used instead.
Crucially, the pseudo-targets are selected on the basis of \emph{relevance} as measured by the Verifier, rather than by IoU with the ground truth.
For each $y \in \mathcal{T}(x)$, temperature-smoothed weights are defined as
\begin{equation}
	w_y = \frac{\exp\bigl(V_\phi(x,y)/\tau\bigr)}{\sum_{y' \in \mathcal{T}(x)} \exp\bigl(V_\phi(x,y')/\tau\bigr)}, \qquad y \in \mathcal{T}(x),
\end{equation}
with temperature $\tau > 0$.
Both $\mathcal{T}(x)$ and $\{w_y\}$ are treated as constants; no gradients flow through the selection step or through $V_\phi$.

Let $\hat{y}_\theta(x)$ denote the Grounder's boundary prediction from the shared temporal head.
The calibration loss takes the form
\begin{equation}
	\mathcal{L}_{\mathrm{cal}}(\theta) = \sum_{y \in \mathcal{T}(x)} w_y \,\ell\bigl(\hat{y}_\theta(x),\, y\bigr),
	\label{eq:calib_loss}
\end{equation}
where $\ell$ is a boundary loss applied to segment endpoints (cross-entropy or KL divergence for discrete heads; SmoothL1 for continuous heads).
Gradients propagate only through $\hat{y}_\theta(x)$.
Employing $K_{\mathrm{cal}}{>}1$ together with a finite $\tau$ distributes the calibration signal across several high-scoring candidates.
This design is intended to reduce over-commitment to a single Verifier-preferred span and to keep the pseudo-target set responsive to improving Grounder proposals over training.

The reward-carrier selection in GRR (Sec.~\ref{subsec:grr}) and the pseudo-target set constructed here serve distinct purposes: the former determines \emph{which trajectory} is reinforced via a single representative segment, whereas the latter governs \emph{how the Grounder's temporal head} is adjusted by averaging over multiple high-confidence spans.
The stop-gradient treatment of $\mathcal{T}(x)$ and $\{w_y\}$ compels the Grounder to improve its own temporal prediction rather than exploiting the Verifier's score surface.

\subsection{Verifier-Based Update Gating}
\label{subsec:gating}

This subsection defines when a sampled group should still contribute to the GRR update.
Drawing on the solver--verifier gap intuition~\cite{sun2025solververifier}, we gate \emph{which groups participate in gradient updates} so that optimization focuses on groups whose best candidate still scores below the current Verifier threshold (Algorithm~\ref{alg:training}).
At each epoch $t$ beyond an initial warmup of $T_{\mathrm{warmup}}$ epochs, a threshold $\tau_t$ is derived from a smoothed average of recent validation Verifier scores, and $A_g$ is set to zero whenever $\bar{v}_g = \max_{y \in \mathcal{C}_\theta(x,c_g)} V_\phi(x,y)$ exceeds $\tau_t$.
The warmup phase prevents premature reliance on Verifier scores before the Grounder has produced a minimally reliable candidate distribution.
Once warmup concludes, zeroing $A_g$ for already-strong groups preserves the sign and scale of advantages for the remaining hard groups, rather than rescaling all trajectories within a batch.

\subsection{Full Objective and Training}
\label{subsec:objective}

This subsection collects the two training terms into the final objective used for optimization.
The complete training objective unifies the policy-gradient and calibration losses:
\begin{equation}
	\mathcal{L}(\theta) = \mathcal{L}_{\mathrm{GRR}}(\theta) + \lambda\,\mathcal{L}_{\mathrm{cal}}(\theta),
	\label{eq:full_loss}
\end{equation}
with $\lambda \ge 0$.
Here $\mathcal{L}_{\mathrm{GRR}}$ acts on language logits through an advantage-weighted NLL, while $\mathcal{L}_{\mathrm{cal}}$ updates the shared temporal head via boundary regression or classification.
Under weak- or no-label conditions, $\mathcal{L}_{\mathrm{cal}}$ provides a differentiable multi-agent signal for temporal refinement; when full supervision is available, it can additionally serve as a regularizer on unlabeled examples.
As made explicit in Algorithm~\ref{alg:training}, Verifier scores are reused at three distinct granularities: selecting a reward carrier within each trajectory, defining the Top-$K_{\mathrm{cal}}$ pseudo-target pool across a group, and gating which groups participate in updates.
This multi-level reuse makes the frozen Verifier a training-time source of structured constraints, all without requiring gradients through $V_\phi$.

\begin{remark}[Relevance vs.\ boundary precision]
	\label{rem:relevance-vs-precision}
	Because the Verifier is trained to emphasize segment-level relevance over boundary precision (Sec.~\ref{subsec:setup}), both the reward in \eqref{eq:reward} and the calibration objective in \eqref{eq:calib_loss} preferentially drive improvements at R@0.3, R@0.5, and mIoU, while providing only a weak gradient signal for strict IoU thresholds such as R@0.7.
	The framework is therefore best suited to tasks in which semantic localization takes precedence over exact boundary placement; narrowing the gap at R@0.7 would necessitate supplementary boundary-level supervision or a Verifier explicitly trained to prioritize boundary precision.
\end{remark}

\section{Experiments}
\label{sec:experiments}

\noindent This section evaluates \model\ and ablates its key components.
The experiments address three questions: whether multi-agent training improves zero-shot transfer over a same-scale baseline, whether the same directional pattern appears across the grounding and long-video QA benchmarks evaluated here, and how the three training components affect the reported results. We therefore report main comparisons on temporal grounding, grounded video QA, and general video QA, followed by component ablations and qualitative error analysis, all under a zero-shot protocol without target-dataset fine-tuning.

\subsection{Experimental Setup}

\paragraph{Datasets}
Evaluation covers benchmarks for temporal grounding, grounded video QA, and general video QA.
For temporal grounding we employ \textbf{Charades-STA}~\cite{gao2017tall} and \textbf{ActivityNet-Captions}~\cite{krishna2017visual}, both of which require localizing intervals from natural-language queries.
Grounded QA is assessed on \textbf{NExT-GQA}~\cite{xiao2024can}, which demands grounding answers to specific video segments.
General video QA is evaluated on \textbf{Video-MME}~\cite{fu2025video} ($\sim$15\,min), \textbf{MLVU}~\cite{zhou2024mlvu} ($\sim$15\,min), and \textbf{LVBench}~\cite{wang2025lvbench} ($\sim$1.1\,h), collectively probing long-video reasoning capabilities.

\paragraph{Evaluation Metrics}
For temporal grounding, we report R@$k$ (recall at IoU threshold $k \in \{0.3, 0.5, 0.7\}$) and mIoU.
For grounded video QA, we report mIoU and Acc@GQA.
For general video QA, we report accuracy metrics as specified by each benchmark.

\paragraph{Implementation Details}
The Grounder is a 2B-parameter multimodal generative model initialized from a pretrained video-language foundation; the Verifier is a separately trained, frozen 2B-parameter model that evaluates candidate segment quality using the supervised data listed in Table~\ref{tab:role_datasets}.
Training is conducted on 8$\times$A100 GPUs with $G=4$ completions sampled per input for GRR, and the reward coefficients are set to $\alpha=1.0$, $\beta_V=0.5$, $\gamma=0.1$, and $\lambda=0.3$ (Eqs.~\eqref{eq:q_reward}--\eqref{eq:reward}, \eqref{eq:full_loss}).
For bootstrapped boundary calibration (Sec.~\ref{subsec:calib}), the Top-$K_{\mathrm{cal}}$ candidates are retained with $K_{\mathrm{cal}}=3$, and a temperature of $\tau=1.0$ governs the Verifier-based weights $w_y$.
The optimizer is AdamW with a learning rate of $2 \times 10^{-5}$ and a batch size of 32.
Regarding Verifier-based gradient-update gating (Algorithm~\ref{alg:training}), the mean Verifier score $m_t$ on validation candidates is computed at each epoch, the gating threshold is set as $\tau_t = \rho \cdot \frac{1}{\min(t,K)}\sum_{t'=t-\min(t,K)+1}^{t} m_{t'}$ with $K=3$ and $\rho=0.95$, and no gating is applied during the first $T_{\mathrm{warmup}}=2$ epochs; beyond this warmup, only groups satisfying $\bar{v}_g < \tau_t$ participate in the GRR update.
All experiments follow a zero-shot protocol on the target datasets, with no task-specific fine-tuning.

\paragraph{Transfer Protocol and Reproducibility}
The transfer protocol is intentionally stringent.
Both Grounder and Verifier are trained exclusively on their respective source datasets, and all reported target-benchmark results are obtained without target-dataset fine-tuning or benchmark-specific adapter tuning.
Verifier triplets are drawn solely from official training splits and remain strictly disjoint from the train/val/test videos used for evaluation.
This separation is essential because the Verifier is reused as a frozen signal across tasks; any leakage between verifier construction and target evaluation would inflate the apparent benefit of multi-agent training.
Table~\ref{tab:hyperparams} consolidates the optimization and selection settings applied consistently throughout the main experiments.

\begin{table}[!t]
	\footnotesize
	\centering
	\caption{\textbf{Training and selection hyperparameters used in the main experiments.} The settings are held fixed across the reported zero-shot transfer results unless otherwise noted.}
	\label{tab:hyperparams}
	\begin{tabularx}{\columnwidth}{l>{\raggedright\arraybackslash}X}
		\toprule
		\textbf{Component} & \textbf{Setting}                                                                            \\
		\midrule
		Group sampling     & $G=4$, reward shortlist $K_r=3$, calibration Top-$K_{\mathrm{cal}}=3$                       \\
		Reward weights     & $\alpha=1.0$, $\beta_V=0.5$, $\gamma=0.1$, $\lambda=0.3$                                    \\
		Calibration        & temperature $\tau=1.0$, verifier-weighted pseudo-targets from the union of group candidates \\
		Gating             & moving-window $K=3$, $\rho=0.95$, warmup $T_{\mathrm{warmup}}=2$                            \\
		Optimization       & AdamW, learning rate $2 \times 10^{-5}$, batch size 32, 8$\times$A100 GPUs                  \\
		Transfer protocol  & zero-shot on Charades-STA, ActivityNet-Captions, NExT-GQA, Video-MME, MLVU, and LVBench     \\
		\bottomrule
	\end{tabularx}
\end{table}

\begin{table*}[!t]
	\centering
	\small
	\caption{Supervised fine-tuning datasets for the Grounder and Verifier.
		The role-wise data configuration follows that of prior work such as VideoMind.
		All verification examples are constructed from the official training splits of the corresponding datasets and are strictly disjoint from the train/val/test splits of the evaluation benchmarks in terms of video or sample identity.}
	\label{tab:role_datasets}
	\begin{tabularx}{\textwidth}{l c >{\raggedright\arraybackslash}X}
		\toprule
		\textbf{Role} & \textbf{\#Samples} & \textbf{Pretraining Datasets}                                                                \\
		\midrule
		Grounder      & 210K               &
		QVHighlights (5K), DiDeMo (33K), TACoS (9K), QuerYD (19K), HiRESTmr (8K), HiRESTstep (4K), CosMo-Cap (87K), InternVid-VTime (54K) \\
		Verifier      & 232K               &
		DiDeMo-Verify (165K), TACoS-Verify (43K), QVHighlights-Verify (24K)                                                               \\
		\bottomrule
	\end{tabularx}
\end{table*}

\begin{table*}[!t]
	\footnotesize
	\setlength{\tabcolsep}{3pt}
	\centering
	\caption{\textbf{Zero-shot temporal grounding on Charades-STA \cite{gao2017tall} and ActivityNet-Captions \cite{krishna2017visual}.} Results are reported under a zero-shot transfer protocol without target-dataset fine-tuning. Gray reference rows denote supervised or fine-tuned methods included for context rather than direct zero-shot comparison. The gray highlight on the final row marks the reported \model\ result.}
	\label{tab:zs_grounding}
	\begin{tabularx}{\textwidth}{l|c|c|>{\centering\arraybackslash}X>{\centering\arraybackslash}X>{\centering\arraybackslash}X>{\centering\arraybackslash}X|>{\centering\arraybackslash}X>{\centering\arraybackslash}X>{\centering\arraybackslash}X>{\centering\arraybackslash}X}
		\toprule
		\multirow{2}{*}{\textbf{Method}}                          & \multirow{2}{*}{\textbf{Size}} & \multirow{2}{*}{\textbf{FT}} & \multicolumn{4}{c|}{\textbf{Charades-STA}} & \multicolumn{4}{c}{\textbf{ActivityNet-Captions}}                                                                                                                                                       \\
		\cmidrule{4-7} \cmidrule{8-11}
		                                                          &                                &                              & R@0.3                                      & R@0.5                                             & R@0.7                  & mIoU                   & R@0.3                  & R@0.5                  & R@0.7                  & mIoU                   \\
		\midrule
		\textcolor{gray}{Moment-DETR~\cite{lei2021detecting}}     & --                             & \textcolor{gray}{\cmark}     & \textcolor{gray}{65.8}                     & \textcolor{gray}{52.1}                            & \textcolor{gray}{30.6} & \textcolor{gray}{45.5} & --                     & --                     & --                     & --                     \\
		\textcolor{gray}{UniVTG~\cite{lin2023univtg}}             & --                             & \textcolor{gray}{\cmark}     & \textcolor{gray}{70.8}                     & \textcolor{gray}{58.1}                            & \textcolor{gray}{35.6} & \textcolor{gray}{50.1} & --                     & --                     & --                     & --                     \\
		\textcolor{gray}{$\rm R^2$-Tuning~\cite{liu2024r2tuning}} & --                             & \textcolor{gray}{\cmark}     & \textcolor{gray}{70.9}                     & \textcolor{gray}{59.8}                            & \textcolor{gray}{37.0} & \textcolor{gray}{50.9} & --                     & --                     & --                     & --                     \\
		\textcolor{gray}{2D-TAN}                                  & --                             & \textcolor{gray}{\cmark}     & --                                         & --                                                & --                     & --                     & \textcolor{gray}{60.4} & \textcolor{gray}{43.4} & \textcolor{gray}{25.0} & \textcolor{gray}{42.5} \\
		\textcolor{gray}{MMN}                                     & --                             & \textcolor{gray}{\cmark}     & --                                         & --                                                & --                     & --                     & \textcolor{gray}{64.5} & \textcolor{gray}{48.2} & \textcolor{gray}{29.4} & \textcolor{gray}{46.6} \\
		\textcolor{gray}{VDI}                                     & --                             & \textcolor{gray}{\cmark}     & --                                         & --                                                & --                     & --                     & \textcolor{gray}{--}   & \textcolor{gray}{48.1} & \textcolor{gray}{28.8} & \textcolor{gray}{--}   \\
		\midrule
		VTimeLLM~\cite{huang2024vtimellm}                         & 13B                            & \xmark                       & 55.3                                       & 34.3                                              & 14.7                   & 34.6                   & --                     & --                     & --                     & --                     \\
		TimeChat~\cite{ren2024timechat}                           & 7B                             & \xmark                       & 51.5                                       & 32.2                                              & 13.4                   & --                     & --                     & --                     & --                     & --                     \\
		Momentor~\cite{qian2024momentor}                          & 7B                             & \xmark                       & 42.6                                       & 26.6                                              & 11.6                   & 28.5                   & 42.9                   & 23.0                   & 12.4                   & 29.3                   \\
		HawkEye~\cite{wang2024hawkeye}                            & 7B                             & \xmark                       & 50.6                                       & 31.4                                              & 14.5                   & 33.7                   & --                     & --                     & --                     & --                     \\
		ChatVTG~\cite{qu2024chatvtg}                              & 7B                             & \xmark                       & 52.7                                       & 33.0                                              & 15.9                   & 34.9                   & 40.7                   & 22.5                   & 9.4                    & 27.2                   \\
		VideoChat-TPO~\cite{li2025tpo}                            & 7B                             & \xmark                       & 58.3                                       & 40.2                                              & 18.4                   & 38.1                   & --                     & --                     & --                     & --                     \\
		E.T.~Chat~\cite{liu2024etbench}                           & 4B                             & \xmark                       & 65.7                                       & 45.9                                              & 20.0                   & 42.3                   & 24.1                   & 12.8                   & 6.1                    & 18.9                   \\
		VideoMind~\cite{liu2025videomind}                         & 2B                             & \xmark                       & 67.6                                       & 51.1                                              & 26.0                   & 45.2                   & 44.0                   & 26.5                   & 12.6                   & 30.1                   \\
		VideoChat~\cite{li2023videochat}                          & 7B                             & \xmark                       & --                                         & --                                                & --                     & --                     & 8.8                    & 3.7                    & 1.5                    & 7.2                    \\
		Video-LLaMA~\cite{videollama2023}                         & 7B                             & \xmark                       & --                                         & --                                                & --                     & --                     & 6.9                    & 2.1                    & 0.8                    & 6.5                    \\
		Video-ChatGPT~\cite{maaz2024videochatgpt}                 & 7B                             & \xmark                       & --                                         & --                                                & --                     & --                     & 26.4                   & 13.6                   & 6.1                    & 18.9                   \\
		Valley~\cite{luo2023valley}                               & 7B                             & \xmark                       & --                                         & --                                                & --                     & --                     & 30.6                   & 13.7                   & 8.1                    & 21.9                   \\
		\midrule
		\rowcolor{gray!15} \model                                 & 2B                             & \xmark                       & \textbf{67.9}                                       & \textbf{51.5}                                              & \textbf{26.0}                   & \textbf{46.1}                   & \textbf{44.7}                   & \textbf{26.8}                   & \textbf{12.4}                   & \textbf{30.5}                   \\
		\bottomrule
	\end{tabularx}
\end{table*}

\subsection{Main Results}

\begin{table*}[!t]
	\footnotesize
	\setlength{\tabcolsep}{2.4pt}
	\centering
	\caption{\textbf{Grounded VideoQA on NExT-GQA \cite{xiao2024can}.} Results are reported without target-dataset fine-tuning. Larger or differently trained systems are included for context, and the primary comparison in the text is to the same-scale 2B baseline. The gray highlight marks the reported \model\ row.}
	\label{nextgqa}
	\begin{tabularx}{\textwidth}{l|c|>{\centering\arraybackslash}X>{\centering\arraybackslash}X>{\centering\arraybackslash}X|>{\centering\arraybackslash}X>{\centering\arraybackslash}X>{\centering\arraybackslash}X|>{\centering\arraybackslash}p{1.2cm}}
		\toprule
		\multirow{2}{*}{\textbf{Method}}               & \multirow{2}{*}{\textbf{Size}} & \multicolumn{3}{c|}{\textbf{IoU}} & \multicolumn{3}{c|}{\textbf{IoP}} & \multirow{2}{*}{\centering\shortstack{\textbf{Acc@}                               \\\textbf{GQA}}} \\
		\cmidrule{3-5} \cmidrule{6-8}
		                                               &                                & R@0.3                             & R@0.5                             & mIoU                                                & R@0.3 & R@0.5 & mIoP        \\
		\midrule
		FrozenBiLM NG+~\cite{frozenbilm2022}           & 890M                           & 13.5                              & 6.1                               & 9.6                                                 & 28.5  & 23.7  & 24.2 & 17.5 \\
		VIOLETv2~\cite{fu2023violetv2}                 & --                             & 4.3                               & 1.3                               & 3.1                                                 & 25.1  & 23.3  & 23.6 & 12.8 \\
		SeViLA~\cite{xiao2023nextgqa_sevila}           & 4B                             & 29.2                              & 13.8                              & 21.7                                                & 34.7  & 22.9  & 29.5 & 16.6 \\
		LangRepo~\cite{langrepo2024}                   & 8$\times$7B                    & --                                & 12.2                              & 18.5                                                & --    & 28.7  & 31.3 & 17.1 \\
		VideoStreaming~\cite{he2024videollm_streaming} & 8.3B                           & --                                & 13.3                              & 19.3                                                & --    & 31.0  & 32.2 & 17.8 \\
		LLoVi~\cite{wang2024llovi}                     & 1.8T                           & --                                & 15.3                              & 20.0                                                & --    & 36.9  & 37.3 & 24.3 \\
		HawkEye~\cite{wang2024hawkeye}                 & 7B                             & 37.0                              & 19.5                              & 25.7                                                & --    & --    & --   & --   \\
		VideoChat-TPO~\cite{li2025tpo}                 & 7B                             & 41.2                              & 23.4                              & 27.7                                                & 47.5  & 32.8  & 35.6 & 25.5 \\
		VideoMind~\cite{liu2025videomind}              & 2B                             & 45.2                              & 23.2                              & 28.6                                                & 51.3  & 32.6  & 36.4 & 25.2 \\
		\midrule
		\rowcolor{gray!15} \model                      & 2B                             & \textbf{46.1}                              & \textbf{23.4}                              & \textbf{28.7}                                                & \textbf{52.2}  & \textbf{32.9}  & \textbf{36.7} & \textbf{25.4} \\
		\bottomrule
	\end{tabularx}
\end{table*}

\begin{figure*}[!t]
	\centering
	\includegraphics[width=\textwidth]{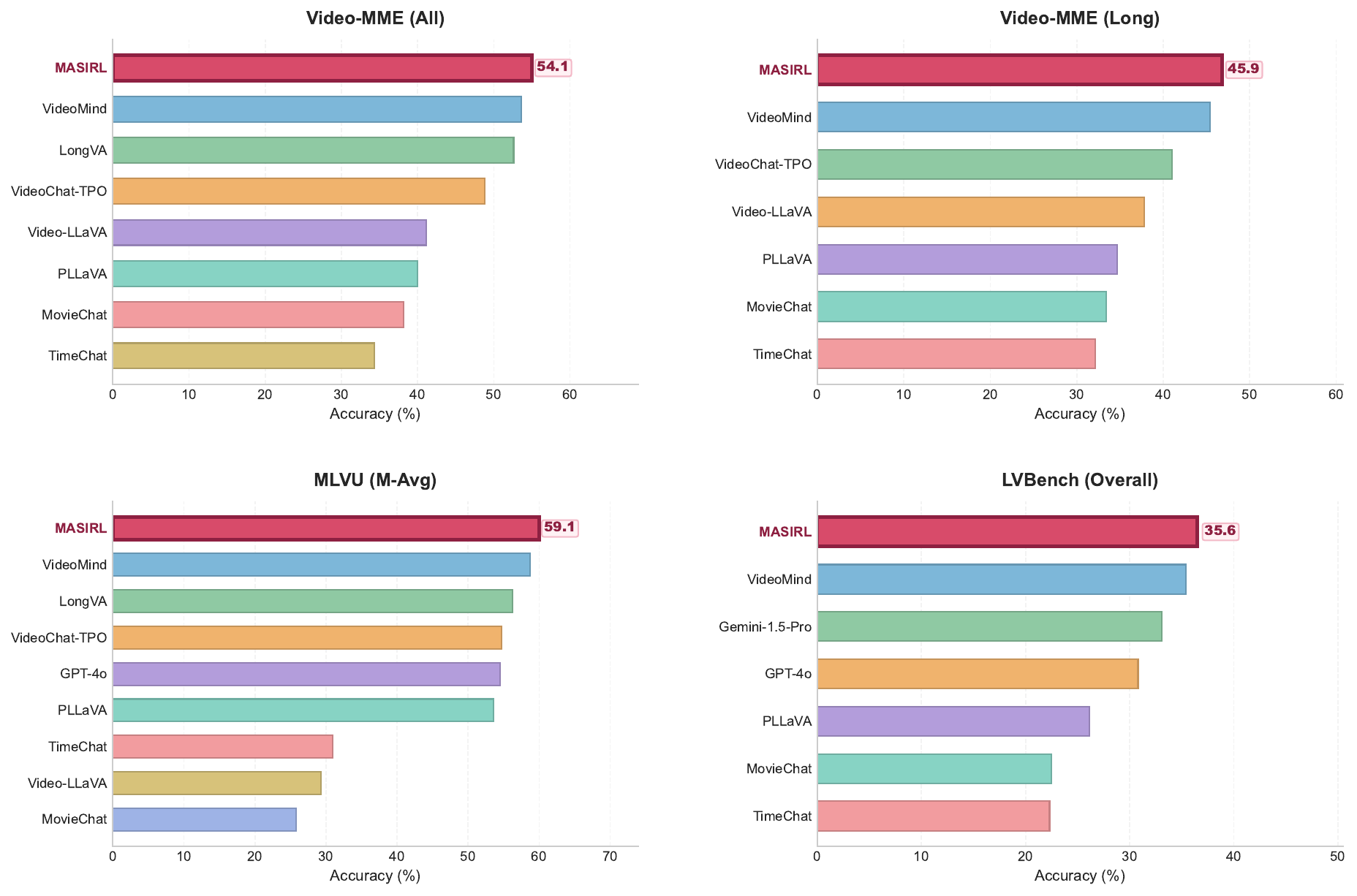}
	\caption{\textbf{General Video QA comparison and gain margin.} Multiple baselines are shown for Video-MME, MLVU \cite{zhou2024mlvu}, and LVBench \cite{wang2025lvbench}; in each panel, methods scoring above \model\ on that benchmark are omitted so that the margins against non-superior baselines remain readable. The \model\ bars are highlighted, while the legend and value labels are placed outside the plotted areas to avoid occlusion.}
	\label{fig:general_qa}
\end{figure*}

\paragraph{Grounded Video Question Answering (NExT-GQA)}
The 2B \model\ attains \textbf{28.7 mIoU} and \textbf{25.4\% Acc@GQA} (Table~\ref{nextgqa}). Relative to the same-scale 2B baseline, VideoMind, the gains are modest but consistent: +0.9 at R@0.3, +0.1 mIoU, and +0.2\% Acc@GQA. IoU- and IoP-based metrics also move upward together (IoP R@0.3: 52.2 vs.\ 51.3; mIoP: 36.7 vs.\ 36.4), so the table does not indicate that the gains come only from selecting broader segments. Comparisons to larger systems are reported for context only. As in the rest of the paper, improvements are clearer at relevance-oriented criteria than at strict boundary thresholds, which is consistent with the Verifier being trained to score segment relevance rather than exact endpoint precision.

\paragraph{General Video Question Answering (Video-MME, MLVU, LVBench)}
On the long-video QA benchmarks summarized in Figure~\ref{fig:general_qa}, \model\ achieves \textbf{59.1\%} on MLVU and \textbf{35.6\%} on LVBench, exceeding the same-scale baseline VideoMind on both benchmarks. For completeness, the corresponding Video-MME results remain \textbf{54.1\%} (All) and \textbf{45.9\%} (Long), also above VideoMind. The margins are again small (+0.2 to +0.5), so the evidence here supports consistency rather than a large step change. Within these three benchmarks and this zero-shot protocol, the direction of improvement is stable across different video lengths and evaluation setups.

\paragraph{Zero-shot Temporal Grounding}
Without target-dataset fine-tuning, \model\ attains \textbf{67.9\% R@0.3}, \textbf{51.5\% R@0.5}, and \textbf{46.1 mIoU} on Charades-STA, together with \textbf{44.7\% R@0.3}, \textbf{26.8\% R@0.5}, and \textbf{30.5 mIoU} on ActivityNet-Captions (Table~\ref{tab:zs_grounding}). The gains over VideoMind are larger on Charades-STA than on ActivityNet-Captions. The table establishes that dataset-wise difference, but it does not by itself determine why the gap is larger on one benchmark than on the other. As anticipated by Remark~\ref{rem:relevance-vs-precision}, the improvements concentrate on mIoU and moderate-overlap recall rather than strict R@0.7, indicating that the current formulation is better at selecting relevant evidence than at refining exact temporal boundaries.

\subsection{Cross-task Consistency and Case Studies}

Across benchmarks, one pattern remains stable. The clearest gains appear on metrics that reward semantic relevance mIoU, R@0.3, R@0.5, and answer accuracy, whereas the strictest boundary criterion, R@0.7, changes little. This behavior is consistent with the way the framework is constructed: both GRR and bootstrapped calibration are driven by a Verifier trained to score segment relevance rather than exact endpoint precision. Positive gains also appear from short-clip grounding benchmarks to the grounded-QA and long-video-QA benchmarks reported here, which supports transfer beyond the verifier's immediate training data within the tested benchmark set.

\subsection{Ablation Studies}
\label{subsec:ablation_main}

We conduct a systematic ablation of the three core components (GRR, bootstrapped calibration, and Verifier-based gating) on NExT-GQA and Video-MME.
Table~\ref{tab:ablation_main} presents the results; a supervised-only baseline that employs neither GRR nor calibration is included for reference.

\begin{table}[!t]
	\centering
	\small
	\caption{\textbf{Component ablation on NExT-GQA and Video-MME.} Each row removes one component from the full \model. ``Sup.\ only'' denotes the supervised-only Grounder without any RL component, included as a reference rather than as a like-for-like ablation of the full training procedure.}
	\label{tab:ablation_main}
	\setlength{\tabcolsep}{3pt}
	\begin{tabular}{@{} l c c c c @{}}  
		\toprule
		\multicolumn{1}{l}{\textbf{Variant}}  
			& \multicolumn{2}{c}{\textbf{NExT-GQA}} & \multicolumn{2}{c}{\textbf{Video-MME}} \\
		\cmidrule(lr){2-3} \cmidrule(lr){4-5}
			& \textbf{mIoU} & \textbf{Acc@GQA} & \textbf{All} & \textbf{Long} \\
		\midrule
		Full \model   & \textbf{28.7} & \textbf{25.4} & \textbf{54.1} & \textbf{45.9} \\
		$-$ GRR       & 28.0          & 24.0          & 52.8          & 44.6          \\
		$-$ Calibration ($\lambda{=}0$) & 28.3 & 24.9 & 53.7 & 45.5 \\
		$-$ Gating    & 28.3          & 25.0          & 53.6          & 45.4          \\
		Sup.\ only    & --            & 23.8          & 52.5          & 44.3          \\
		\bottomrule
	\end{tabular}
\end{table}

\paragraph{Role of GRR}
Removing GRR incurs the largest single-component drop: $-0.7$ mIoU and $-1.4$\% Acc@GQA on NExT-GQA, together with $-1.3$\% on both Video-MME splits.
The larger drop on Acc@GQA than on mIoU ($-1.4$\% vs.\ $-0.7$) is consistent with GRR affecting answer generation as well as evidence selection, though the ablation alone does not isolate the full mechanism.
At minimum, Table~\ref{tab:ablation_main} shows that removing GRR weakens both the grounding and QA metrics reported here.
This pattern is compatible with the intended role of GRR: applying relative credit assignment at the trajectory level rather than updating the temporal and language parts of the model only through separate objectives.

\paragraph{Role of calibration}
Setting $\lambda{=}0$ (no calibration) yields 28.3 mIoU and 24.9\% Acc@GQA on NExT-GQA, corresponding to drops of $-0.4$ mIoU and $-0.5$\% Acc@GQA relative to the full model.
The removal affects localization and QA only moderately, with a somewhat clearer change on NExT-GQA mIoU than on the Video-MME accuracies shown in the same table.
This pattern is consistent with calibration acting primarily through the temporal head, since the calibration loss regresses boundary predictions toward Verifier-weighted pseudo-targets whereas GRR acts through token-level NLL.
The accompanying 0.5\% drop in Acc@GQA indicates that the calibration term is not neutral for QA behavior, although Table~\ref{tab:ablation_main} does not by itself establish the precise pathway for that effect.

\paragraph{Role of gating}
Removing gating reduces NExT-GQA performance to 28.3 mIoU / 25.0\% Acc@GQA and Video-MME to 53.6\% / 45.4\%.
Although the magnitude of drop is comparable to that of removing calibration, the two ablations remove different training components.
Under the formulation in Sec.~\ref{subsec:gating}, gating excludes groups whose best candidate already exceeds the current threshold, whereas removing gating allows every group to contribute to every GRR update.
The table therefore supports the claim that this selection step is helpful in the reported setting, but it does not on its own establish a stronger statement about when in training the benefit is largest.

\begin{figure}[!t]
	\centering
	\begin{tikzpicture}
		\begin{axis}[
				width=\columnwidth,
				height=4.7cm,
				xmin=-0.02,
				xmax=0.52,
				ymin=24.6,
				ymax=28.9,
				xlabel={$\lambda$},
				ylabel={Score (\%)},
				title={Sensitivity to $\lambda$},
				legend style={font=\scriptsize, draw=none, at={(0.5,1.18)}, anchor=south, legend columns=2},
				tick label style={font=\footnotesize},
				label style={font=\footnotesize},
				title style={font=\footnotesize},
				grid=both,
				grid style={dashed,gray!30}
			]
			\addplot[
				color=blue!70!black,
				mark=*,
				line width=1pt
			] coordinates {(0,28.3) (0.1,28.5) (0.3,28.7) (0.5,28.4)};
			\addlegendentry{mIoU}
			\addplot[
				color=orange!85!black,
				mark=square*,
				line width=1pt
			] coordinates {(0,24.9) (0.1,25.1) (0.3,25.4) (0.5,25.0)};
			\addlegendentry{Acc@GQA}
			\addplot[
				color=gray!70,
				densely dashed,
				forget plot
			] coordinates {(0.3,24.6) (0.3,28.9)};
		\end{axis}
	\end{tikzpicture}
	\vspace{0.6em}
	\begin{tikzpicture}
		\begin{axis}[
				width=\columnwidth,
				height=4.7cm,
				xmin=1.5,
				xmax=8.5,
				ymin=24.6,
				ymax=28.9,
				xlabel={$G$},
				ylabel={Score (\%)},
				title={Sensitivity to group size $G$},
				legend style={font=\scriptsize, draw=none, at={(0.5,1.18)}, anchor=south, legend columns=2},
				xtick={2,4,8},
				tick label style={font=\footnotesize},
				label style={font=\footnotesize},
				title style={font=\footnotesize},
				grid=both,
				grid style={dashed,gray!30}
			]
			\addplot[
				color=blue!70!black,
				mark=*,
				line width=1pt
			] coordinates {(2,28.2) (4,28.7) (8,28.6)};
			\addlegendentry{mIoU}
			\addplot[
				color=orange!85!black,
				mark=square*,
				line width=1pt
			] coordinates {(2,24.8) (4,25.4) (8,25.3)};
			\addlegendentry{Acc@GQA}
			\addplot[
				color=gray!70,
				densely dashed,
				forget plot
			] coordinates {(4,24.6) (4,28.9)};
		\end{axis}
	\end{tikzpicture}
	\caption{\textbf{Sensitivity to $\lambda$ and $G$ on NExT-GQA in figure form.} Line plots make the local trend around the selected setting easier to compare than the original table. The dashed vertical guide marks the main setting used in the experiments: $\lambda{=}0.3$ and $G{=}4$.}
	\label{fig:sensitivity}
\end{figure}
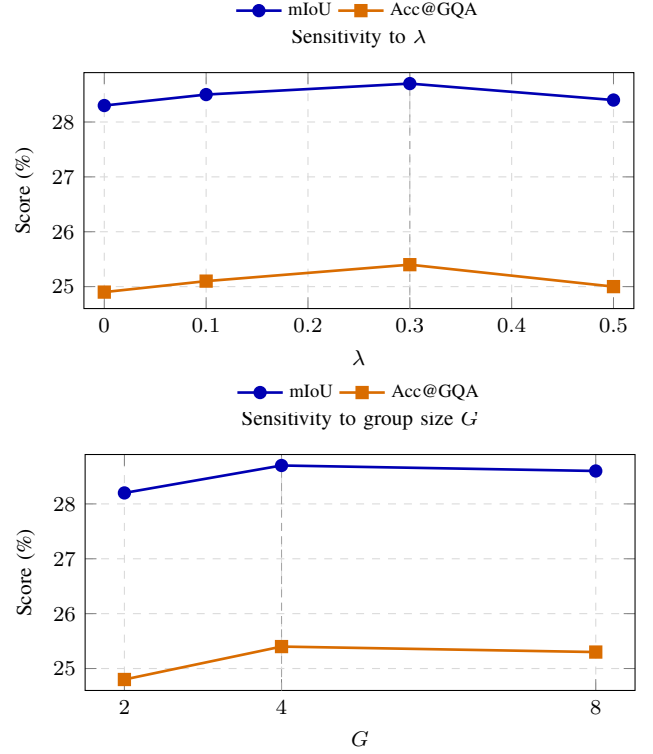

\paragraph{Sensitivity to $\lambda$ and $G$}
Figure~\ref{fig:sensitivity} reports the sensitivity of mIoU and Acc@GQA on NExT-GQA to the calibration weight $\lambda$ and the group size $G$.
For $\lambda$, 0.3 is the best among the tested values: smaller values score lower, whereas $\lambda{=}0.5$ reduces Acc@GQA by 0.4\% relative to the full setting.
The non-monotonic pattern is consistent with the two losses affecting different parts of training (sequence-level policy vs.\ boundary regression), although Figure~\ref{fig:sensitivity} only supports that point within the tested grid.
For $G$, the transition from $G{=}2$ to $G{=}4$ improves both metrics (+0.5 mIoU, +0.6\% Acc@GQA).
Increasing further to $G{=}8$ changes the results only slightly ($-0.1$ mIoU, $-0.1$\% Acc@GQA) at higher sampling cost, so the reported trade-off favors $G{=}4$ within this sweep.

\paragraph{Reward-carrier Selection}
In the main experiments, reward carriers are selected via Verifier ranking with a Top-$K_r$ shortlist ($K_r{=}3$).
Table~\ref{tab:reward_carrier_ablation} compares three selection strategies on zero-shot temporal grounding (averaged over Charades-STA and ActivityNet-Captions).
\begin{table}[!t]
	\centering
	\small
	\caption{\textbf{Reward-carrier selection strategies.}
		Results are averages over Charades-STA and ActivityNet-Captions under the same zero-shot temporal-grounding setting used in the main comparison.}
	\label{tab:reward_carrier_ablation}
	\begin{tabular}{lcc}
		\toprule
		Selection strategy                & Avg R@0.5     & Avg mIoU      \\
		\midrule
		First-valid segment               & 38.5          & 37.8          \\
		Top-1 by Verifier score           & 38.9          & 38.1          \\
		Top-$K_r$ by Verifier ($K_r{=}3$) & \textbf{39.3} & \textbf{38.4} \\
		\bottomrule
	\end{tabular}
\end{table}
All three strategies yield comparable results, yet Top-$K_r$ holds a marginal advantage (+0.8 R@0.5, +0.6 mIoU over first-valid).
Within this comparison, the narrow gap indicates that reward-carrier selection has a smaller effect than the larger component drops reported in Table~\ref{tab:ablation_main}.
The Top-$K_r$ variant remains the strongest of the three tested strategies, which is consistent with using the Verifier to exclude some low-ranked candidates before reward computation.

\subsection{Qualitative Analysis}

\begin{figure*}[!t]
	\centering
	\includegraphics[width=0.9\textwidth]{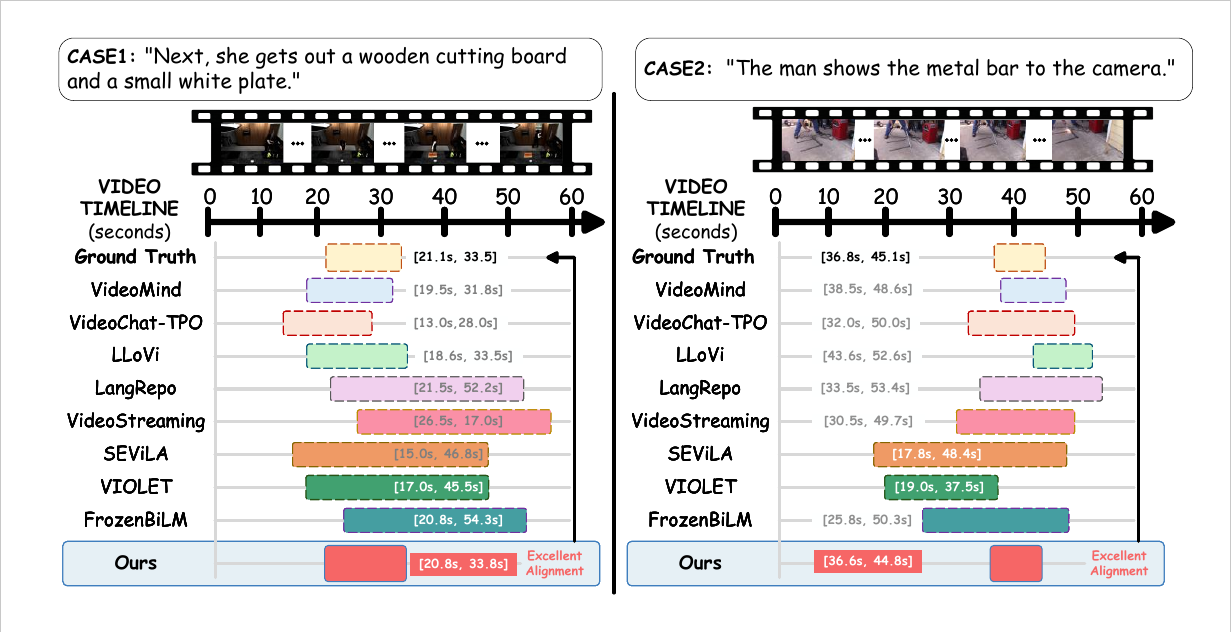}
	\caption{Illustrative qualitative comparison of temporal grounding results on representative video queries.
		\textbf{Left:} Case 1: ``Next, she gets out a wooden cutting board and a small white plate.''
		\textbf{Right:} Case 2: ``The man shows the metal bar to the camera.''
		For each case, the predicted temporal intervals of \model\ and multiple baselines are shown aligned with the video timeline.
		Ground truth intervals are indicated by dashed boxes. The figure is intended as qualitative support for the quantitative trends discussed in the text, not as a stand-alone causal test.}
	\label{fig:qualitative}
\end{figure*}

Figure~\ref{fig:qualitative} presents two representative cases illustrating qualitative behavior associated with the quantitative trends above.
We analyze each case to connect qualitative behavior to quantitative patterns from the main results.

\paragraph{Case 1: ``Next, she gets out a wooden cutting board and a small white plate.''}
This Charades-STA query describes a brief, well-defined action embedded within a longer cooking sequence.
The ground truth spans a short interval during which a person reaches for and places both items.
Several baselines (e.g., TimeChat, Momentor) predict intervals that begin too early, capturing a preceding cabinet-opening action that is semantically related yet distinct.
VideoMind produces a tighter prediction but still includes ${\sim}2$\,s of lead-in context.
By contrast, \model\ returns an interval that closely matches the ground truth, excluding the cabinet-opening entirely.

This case is qualitatively consistent with the relevance-oriented gains reported earlier.
In the example, the baselines overlap the target action but include more lead-in context, whereas \model\ is visually closer to the annotated interval.
One plausible interpretation is that Verifier scoring, GRR, and calibration together favor trajectories that place more weight on the action-bearing portion of the sequence, but the figure is illustrative rather than a direct causal test.

\paragraph{Case 2: ``The man shows the metal bar to the camera.''}
This ActivityNet-Captions query describes a demonstrative action within a longer instructional video.
The central challenge is that the man handles a metal bar throughout much of the video yet only briefly holds it up to the camera.
Baselines tend to predict a broad interval covering the entire handling sequence, which is semantically relevant but temporally imprecise.
\model\ narrows the prediction to the specific ``showing'' gesture, achieving higher IoU with the ground truth.

This example likewise matches the broader pattern that the model can separate a target action from nearby semantically related context.
The figure is compatible with the intended use of the Verifier as a preference signal over candidate spans, but it should be read as an illustrative case rather than proof of a specific internal decision rule.

\paragraph{Common failure patterns}
In a recurring failure mode, the Grounder retrieves a segment containing the correct action but extending $3$--$5$\,s beyond the annotated endpoint.
For instance, on a query about ``the person closes the door,'' \model\ correctly identifies the closing action but retains several frames of the person walking away. Although these frames are semantically compatible with the query context, they fall outside the annotated boundaries.
The Verifier scores this extended segment as highly relevant (the door-closing is clearly present), so neither GRR nor calibration receives a signal to trim the trailing frames.
This failure is consistent with a limitation of relevance-based scoring: when post-action context remains semantically coherent with the query, the current training signal may provide little pressure to trim the trailing frames.

\subsection{Boundary Preference and Error Taxonomy}

Three recurring failure patterns, all stemming from the relevance oriented bias inherent in the Verifier's objective, are evident in the qualitative analysis. First, early start errors occur when a segment initiates before the action bearing window, typically triggered by semantically compatible preparatory context. The Verifier fails to penalize this extension because the core action remains present, resulting in a decline in R@0.7 while R@0.3 and mIoU remain unaffected. Second, late end errors reflect a symmetric issue wherein trailing aftermath frames are retained; the metric impact is analogous, mildly reducing mIoU without affecting the lenient R@0.3 threshold. Third, context over inclusion arises from ambiguous queries or shared visual features across sub actions, leading the model to select an entire sub activity rather than the specific target action. While R@0.3 may persist, R@0.5, R@0.7, and mIoU deteriorate notably.

Critically, these error modes are interdependent and stem from the Verifier's optimization target: distinguishing broadly relevant evidence from irrelevant content, rather than adjudicating fine grained boundary discrepancies. Consequently, any segment containing the target action receives a high score irrespective of extraneous preparatory or trailing context. Bridging the gap between relevance and precision would likely require modifying the training regime to incorporate boundary sensitive negatives. Specifically, this would involve paired segments differing only in temporal bounds, where the tighter annotation is preferred. Although such a boundary aware Verifier remains a hypothesis for future investigation, its successful implementation could sharpen pseudo targets for the calibration loss and yield more discriminative rewards for GRR. In summary, the evidence substantiates consistent and same scale improvements on relevance metrics through the integration of the Verifier signal in GRR, calibration, and gating. The primary residual challenge lies in contexts where neighboring temporal spans are semantically indistinguishable, thereby limiting the guidance provided by the current relevance oriented signal for precise boundary demarcation.

\section{Limitations}
\label{sec:limitations}

\noindent Several limitations follow directly from the evidence in the main results and qualitative analysis. First, the training signal depends on a frozen Verifier trained for segment-level relevance rather than exact boundary discrimination. That dependence helps explain the observed pattern that gains are clearer on mIoU and moderate-overlap recall than on strict thresholds such as R@0.7, and it is also consistent with the early-start, late-end, and context over-inclusion errors discussed in Secs.~\ref{sec:experiments}. When multiple neighboring spans all contain the target action, the current Verifier provides limited pressure to prefer the tighter boundary. Second, the gains over the strongest same-scale 2B baseline are positive but small across the reported benchmarks, so the paper supports consistency within this zero-shot benchmark set more than a large margin of superiority. Third, the training loop is more expensive than supervised fine-tuning alone because each update combines grouped trajectory sampling with repeated Verifier scoring. Finally, the study does not test alternative Verifier calibrations or a boundary-aware Verifier objective, so any claim that those changes would improve strict-threshold metrics remains a hypothesis rather than a demonstrated result.

\section{Conclusion}
\label{sec:conclusion}

\noindent This paper studies a narrow question: whether a frozen Verifier can help train a video Grounder, rather than serving only as an inference-time selector. Within the tested zero-shot transfer setting, the results support a limited but consistent yes. The key value of the framework lies in how the Grounder and Verifier interact during training: the Grounder proposes candidate evidence spans and answer-bearing trajectories, and frozen verification turns those candidates into a stable preference signal that is reused in GRR, calibration, and gating. Across grounded QA, temporal grounding, and long-video QA, this design improves a same-scale 2B baseline with the clearest gains on mIoU, moderate-overlap recall, and related relevance-oriented measures. At the same time, the evidence also bounds the claim. The gains remain modest, strict boundary precision improves less, and the experiments do not establish broader transfer or practical impact beyond the evaluated benchmarks.

This interpretation therefore remains cautious. The method is better viewed as a relevance-oriented training strategy for video evidence selection than as a complete solution to fine-grained temporal localization, and a natural next step is to make frozen verification more boundary-aware so it can better distinguish tight spans from over-inclusive ones.
\bibliographystyle{IEEEtran}
\bibliography{example_paper}

\section*{Author Biography}

\begin{IEEEbiography}[{\includegraphics[width=1in,height=1.25in,clip,keepaspectratio]{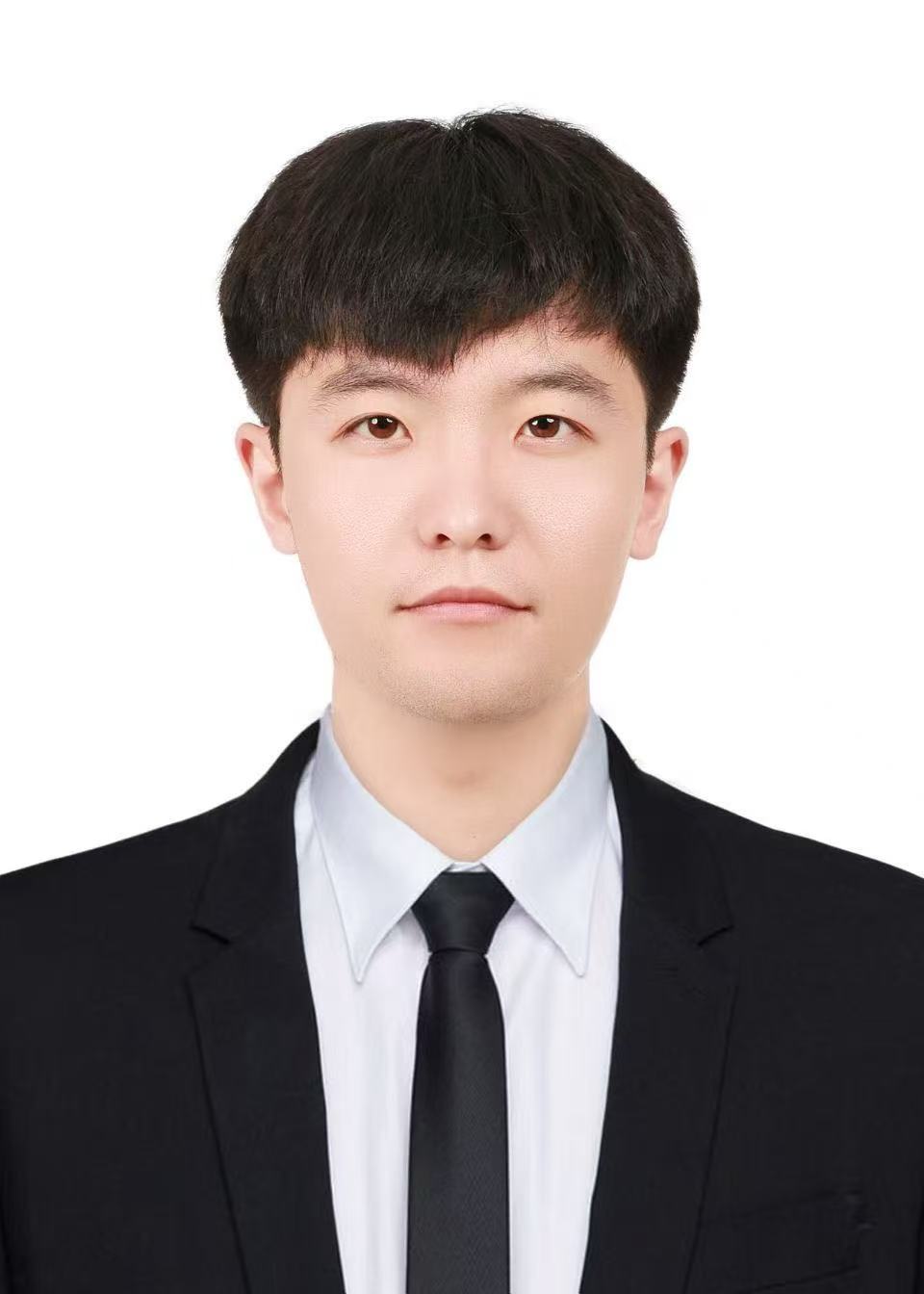}}]
{Mingwen Zhang} is currently pursuing the Ph.D. degree with the School of Information Science and Engineering, Lanzhou University, China. His current research interests include affective computing, with a particular focus on medical image segmentation and pattern recognition.
\end{IEEEbiography}
 
\begin{IEEEbiography}[{\includegraphics[width=1in,height=1.25in,clip,keepaspectratio]{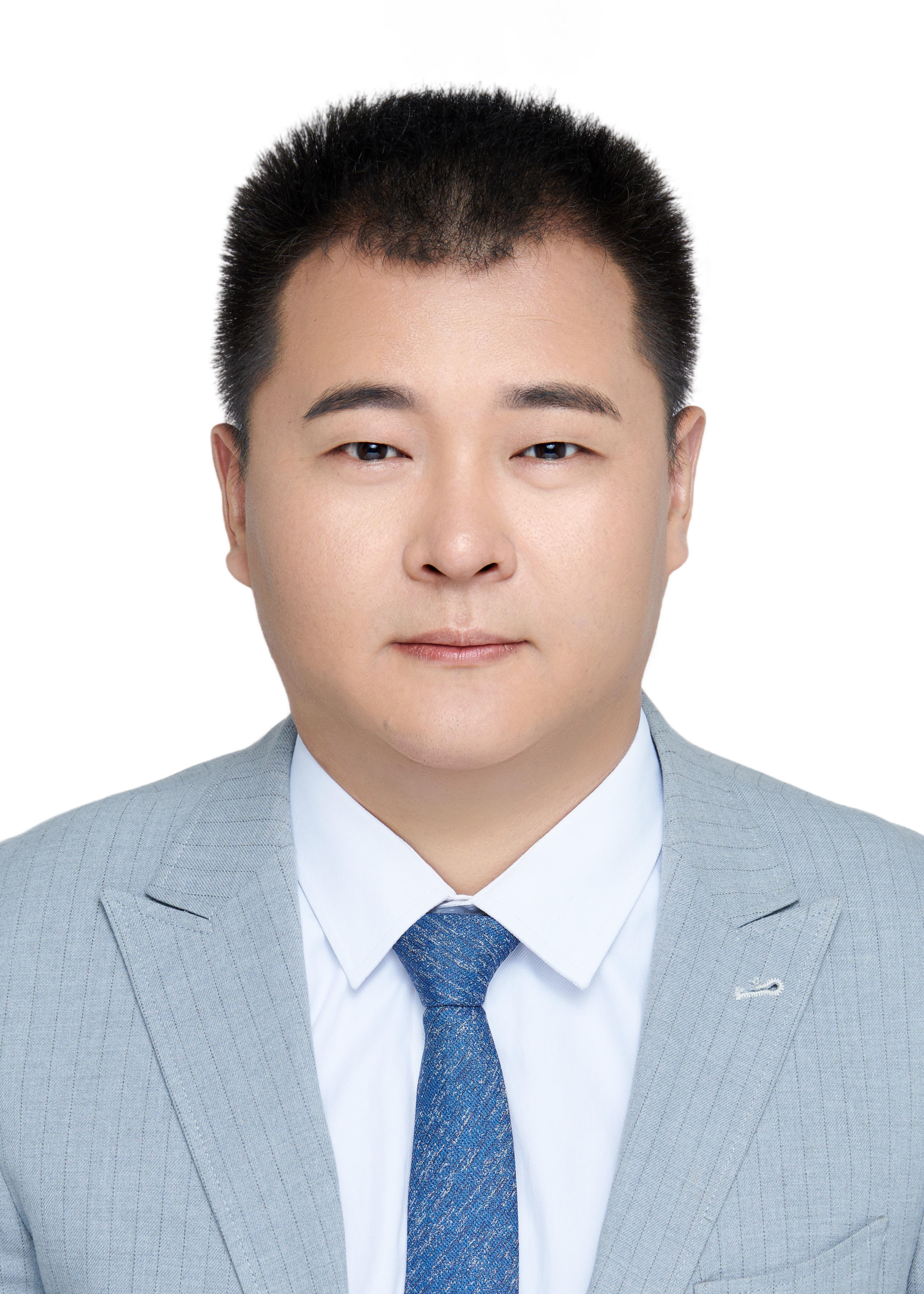}}]
{Jisheng Dang} received the Ph.D. degree from Sun Yat-sen University, China, in 2025, advised by prof. Jianhuang Lai and prof. Huicheng Zheng.
He worked as a research fellow at the NExT++ laboratory of the NationalT University of Singapore  advised by prof. Tat-Seng Chua. He is now a tenured associate professor at the School of Information Science and Engineering, Lanzhou University. His research interests include multimodal learning, video understanding, and embodied  intelligence. 
He has published several papers as the first author in major journals and conferences including IEEE TIP/TNNLS/TITS/IJCAI/AAAI. He served as a reviewer at some major journals and conferences like IEEE TPAMI, ICML, NIPS, ICLR, IEEE TIP, CVPR, IJCAI, ACM MM, AAAI, IEEE TMM, IEEE TCSVT, ACM TOMM.
\vspace{10 mm}
\end{IEEEbiography}

\begin{IEEEbiography}[{\includegraphics[width=1in,height=1.25in,clip,keepaspectratio]{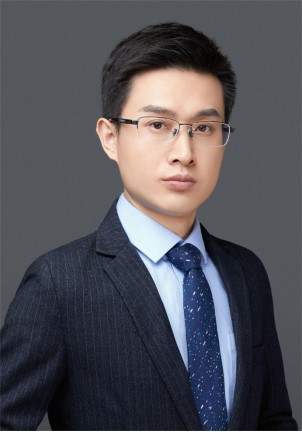}}]{Minqiang Yang}(Member, IEEE) received his Ph.D. degree in computer science from Lanzhou University. He is currently an associate professor with the Gansu Provincial Key Laboratory of Wearable Computing, School of Information Science and Engineering, Lanzhou University. His current research interests include affective computing, image processing, human-computer interaction, pervasive computing, particularly focusing on intelligent diagnostic and therapeutic technologies for affective disorders. He has published over 30 papers in prestigious journals and conferences, including PIEEE, IEEE TKDE, IEEE WCM, IEEE TCSVT, IEEE IoTJ, IEEE JBHI, IEEE TCSS, and IEEE BIBM, and has been granted over 10 patents.
\vspace{10 mm}
\end{IEEEbiography}

\begin{IEEEbiography}[{\includegraphics[width=1in,height=1.25in,clip,keepaspectratio]{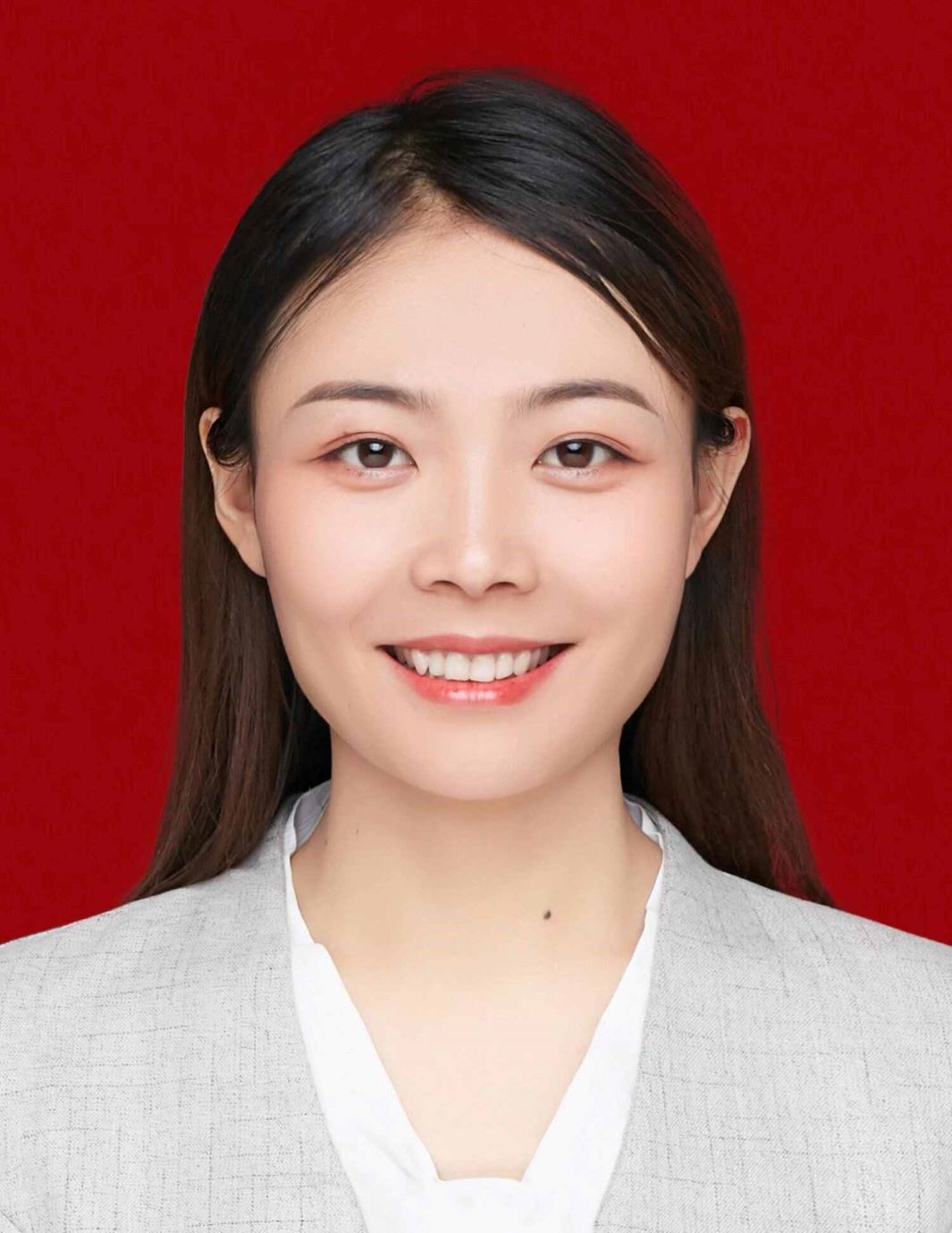}}]{Bimei Wang} (Member, IEEE) received her Ph.D. degree from Jinan University, under the supervision of Prof. Jian Weng, a recipient of the National Young Talent Program. From June 2024, she was a visiting Ph.D. student at the School of Computing, National University of Singapore, guided by Prof. Ee-Chien Chang. Since 2026, she has been an associate professor of the School of Information Science and Engineering at Lanzhou University. Her research interests focus on large language models and their security. She has published multiple papers as both the first author and corresponding author. Additionally, she has filed more than ten Chinese patents and one U.S. patent, and has served as a reviewer for prominent international conferences, including AAAI and ICME.
\vspace{10 mm}
\end{IEEEbiography}

\begin{IEEEbiography}[{\includegraphics[width=1in,height=1.25in,clip,keepaspectratio]{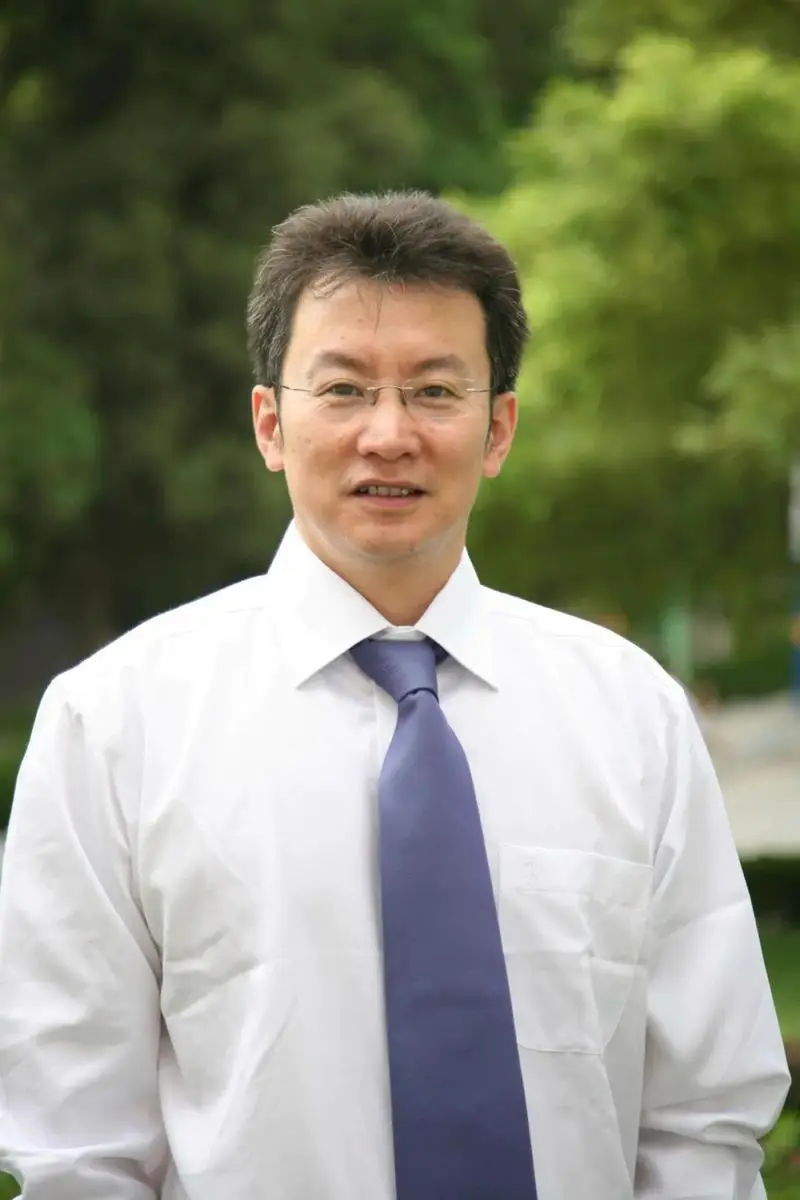}}]
{Bin Hu}(Fellow, IEEE) was a recipient of many research awards, including the 2014 China Overseas Innovation Talent Award, the 2016 Chinese Ministry of Education Technology Invention Award, the 2018 Chinese National Technology Invention Award, and the 2019 WIPO-CNIPA Award for Chinese Outstanding Patented Invention.
He is also the TC Co-Chair of computational psychophysiology in the IEEE Systems, Man, and Cybernetics Society (SMC), the TC Co-Chair of cognitive computing in IEEE SMC, and the Vice-Chair of the TC 9.1. Economic, Business, and Financial Systems on Social Media at the International Federation
of Automatic Control (IFAC). He is also a Member-at-Large of the ACM China Council and the Vice-Chair of the China Committee of the International Society for Social Neuroscience. He serves as the Editor-in-Chief for IEEE TRANSACTIONS ON COMPUTATIONAL SOCIAL SYSTEMS and an Associate Editor for IEEE TRANSACTIONS ON AFFECTIVE COMPUTING.
\end{IEEEbiography}

\begin{IEEEbiography}[{\includegraphics[width=1.1in,height=1.3in,clip,keepaspectratio]{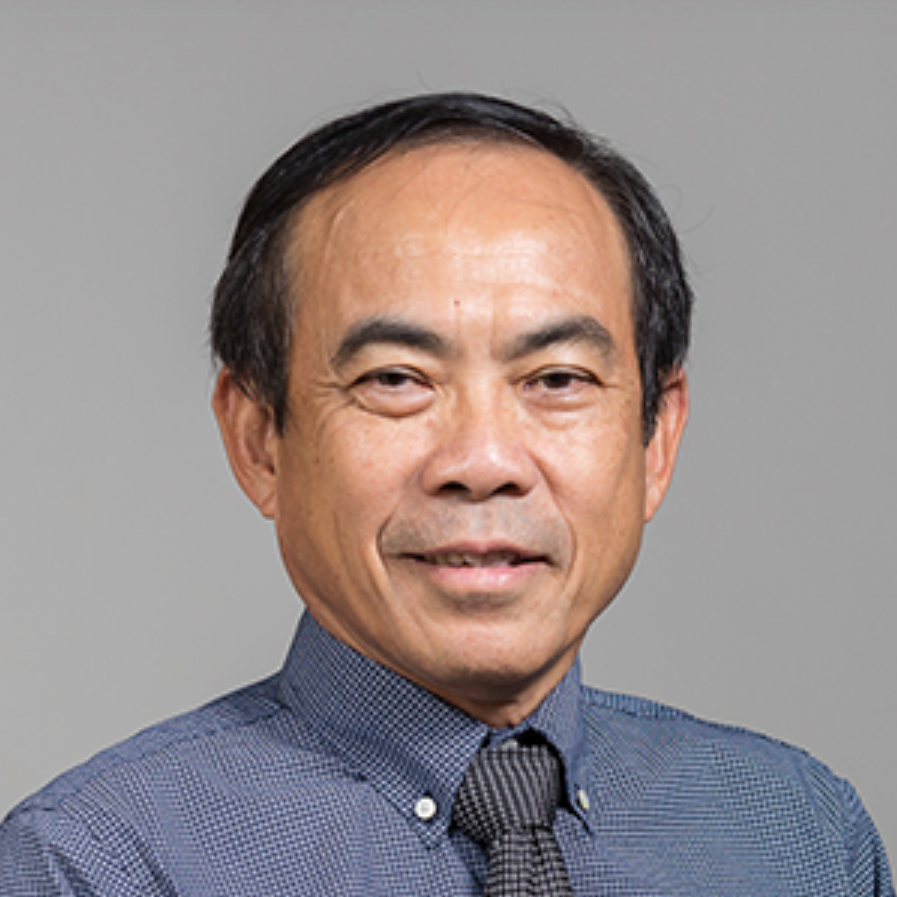}}]
{Tat-Seng Chua} received the Ph.D. degree from the University of Leeds, U.K. He is the KITHCT chair professor with the School of Computing, National University of Singapore, where he was the acting and founding dean of the School from 1998 to 2000. He is the co-director of NExT, a joint center between NUS and Tsinghua University, to develop technologies for live social media search. He is the 2015 winner of the prestigious ACM SIGMM Award. He is the chair of Steering Committee of the ACM International Conference on Multimedia Retrieval (ICMR) and Multimedia Modeling (MMM) conference series. He is also the general co-chair of ACM Multimedia 2005, ACM CIVR (now ACM ICMR) 2005, ACM SIGIR 2008, and ACM Web Science 2015. He serves on the editorial boards of four international journals. He is the co-founder of two technology startups in Singapore and a Fellow of the Singapore Academy of Sciences, with 103,316 citations on Google Scholar.
\end{IEEEbiography}

\end{document}